\documentclass{article}
\pdfoutput=1
\usepackage{hyperref}
\hypersetup{bookmarks=false,linkcolor=blue,urlcolor=blue,colorlinks,citecolor=blue}
\usepackage{arxiv}
\usepackage[utf8]{inputenc} %
\usepackage[T1]{fontenc}    %
\usepackage{graphicx}
\usepackage[titletoc,title]{appendix}
\usepackage[
backend=biber,
style=ieee,
natbib=true,
citestyle=numeric
]{biblatex}
\title{The Cost of Training NLP Models\\ \large A Concise Overview}
\shorttitle{The Cost of Training NLP Models: A Concise Overview}

\author{Or Sharir\\
AI21 Labs\\
\texttt{ors@ai21.com}
\And
Barak Peleg \\
AI21 Labs\\
\texttt{barakp@ai21.com}
\And
Yoav Shoham\\
AI21 Labs\\
\texttt{yoavs@ai21.com}
}
\date{April 2020}

\begin{document}

\maketitle

\begin{abstract}
    We review the cost of training large-scale language models, and the drivers of these costs. The intended audience includes engineers and scientists budgeting their
    model-training experiments, as well as non-practitioners trying to make sense
    of the economics of modern-day Natural Language Processing~(NLP).\footnote{We thank Barak Lenz, Shai Shalev-Shwartz and other members of AI21 Labs, as
    well as Jack Clark, Jeff Dean, Deep Ganguli, Chris Re, Sebastian Ruder and Lior Wolf, who generously  
    commented on previous drafts. Further comments on the document are welcome, and
    the document will be updated as appropriate. Note: While the comments of our
    colleagues from other organizations greatly improved the document, they were
    not representing their organizations, did not share any proprietary information,
    and may not necessarily agree with everything written here.}
\end{abstract}

\section{Costs: Not for the faint hearted}

The cost of floating-point operations (FLOPs), the basic Neural Network~(NN) operation, has been
decreasing. For example, Google reported~\citep{GoogleTPU} a 38\% cost decrease
in ResNet-50 training costs\footnote{It also reported a dramatic 27$\times$ decrease
in training time. While training time is not our focus, it is relevant indirectly:
Compressed time makes it realistic to train larger models, which costs more.}. This was achieved
with optimized hardware (moving from GPUs to TPUs) coupled with framework-level optimizations,
exploiting parallelism opportunities. This kind of cost reduction isn’t an isolated
occurrence~--~we’re seeing the costs of training large models fall as hardware innovations and
training techniques improve. Despite this, overall costs have increased, and can run into the
millions. We’ll explain why this is occurring and what factors play a significant role in the
costs of training\footnote{There is a whole other discussion to be had on the costs of NLP models at
inference time. These are quite related to the training costs, but deserve a separate discussion. In particular, the inference phase 
allows for post-training model optimizations, for example via model
distillation~\citep{sanh2019distilbert,jiao2019tinybert}. This discussion is beyond the scope of this
article.} NLP models.

Just how much does it cost to train a model? Two correct answers are ``depends'' and
``a lot''. More quantitatively, here are current ballpark list-price costs of training differently
sized BERT~\citep{devlin2019Bert} models on the Wikipedia and Book corpora~(15 GB). For each setting we report two numbers - the cost of one training run, and a typical fully-loaded cost (see discussion of "hidden costs" below) with hyper-parameter tuning and multiple runs per setting (here we look at a somewhat modest upper bound of two configurations and ten runs per configuration).\footnote{The
following figures are based on internal AI21 Labs data. They can be somewhat lower due to discounts, or using preemptible
versions of the system. The figures also assume the use of cloud solutions such as
GCP or AWS, and on-premise implementations are sometimes cheaper. Still, the figures
provide a general sense of the costs. }
\begin{itemize}
    \item \$2.5k - \$50k (110 million parameter model)
    \item \$10k - \$200k (340 million parameter model)
    \item \$80k - \$1.6m (1.5 billion parameter model)
\end{itemize}

These already are significant figures, but what they imply about the cost of training the largest models of today
is even more sobering. Exact figures are proprietary information of the specific companies,
but one can make educated guesses. For example, based on information released by
Google, we estimate that, at list-price, training the 11B-parameter variant\footnote{
With context lengths of 512 for both encoding and decoding, 128 attention heads, and
65k-dimensional feed-forward layers.} of T5~\citep{2019t5}
 cost well above \$1.3 million for a single run. Assuming 2-3 runs of the large
model and hundreds of the small ones, the (list-)price tag for the entire project may have been \$10 million\footnote{These \$ figures come with substantial error bars,
but we believe they are in the right ballpark.}.

Not many companies~--~certainly not many startups~--~can afford this cost. Some
argue that this is not a severe issue; let the Googles of the world pre-train and
publish the large language models, and let the rest of the world fine-tune them (a much
cheaper endeavor) to specific tasks. Others (e.g., \citet{CloudAndInnovation}) are not as sanguine. 

\section{Cost Drivers: Size Matters}

We are not aware of a formula that tells you how many FLOPs are needed in a given
NLP setting to achieve a given performance\footnote{
It is worth noting the work of \citep{kaplan2020scaling}, which
analyzes the impact of various variables, including
model size and amount of compute, on performance, as measured by perplexity.
Although the paper does not directly address the question we are after, the
methodology it offers may provide useful hints. Other relevant papers include~\citep{dodge-etal-2019-show,li2020train,rosenfeld2020a}.
}.
However, there are several variables that impact this number, all of which have
increased dramatically in the past few years, far surpassing the once-deemed
``massive'' vision-focused ML models.\footnote{Although computer vision is not our focus here, the contrast with NLP is striking, and we discuss it briefly in Appendix~\ref{app:NLPvsCV}.} 

Here are some of the relevant variables, which fall into three categories: (a) size of dataset, (b) model size (we use the number of parameters as a proxy), and (c) training volume (we use as proxy the total number of tokens processed during pre-training). The top row applies to all models, and the bottom row zooms in on transformer-based models.

\vspace{0.3cm}

\includegraphics[width=\linewidth]{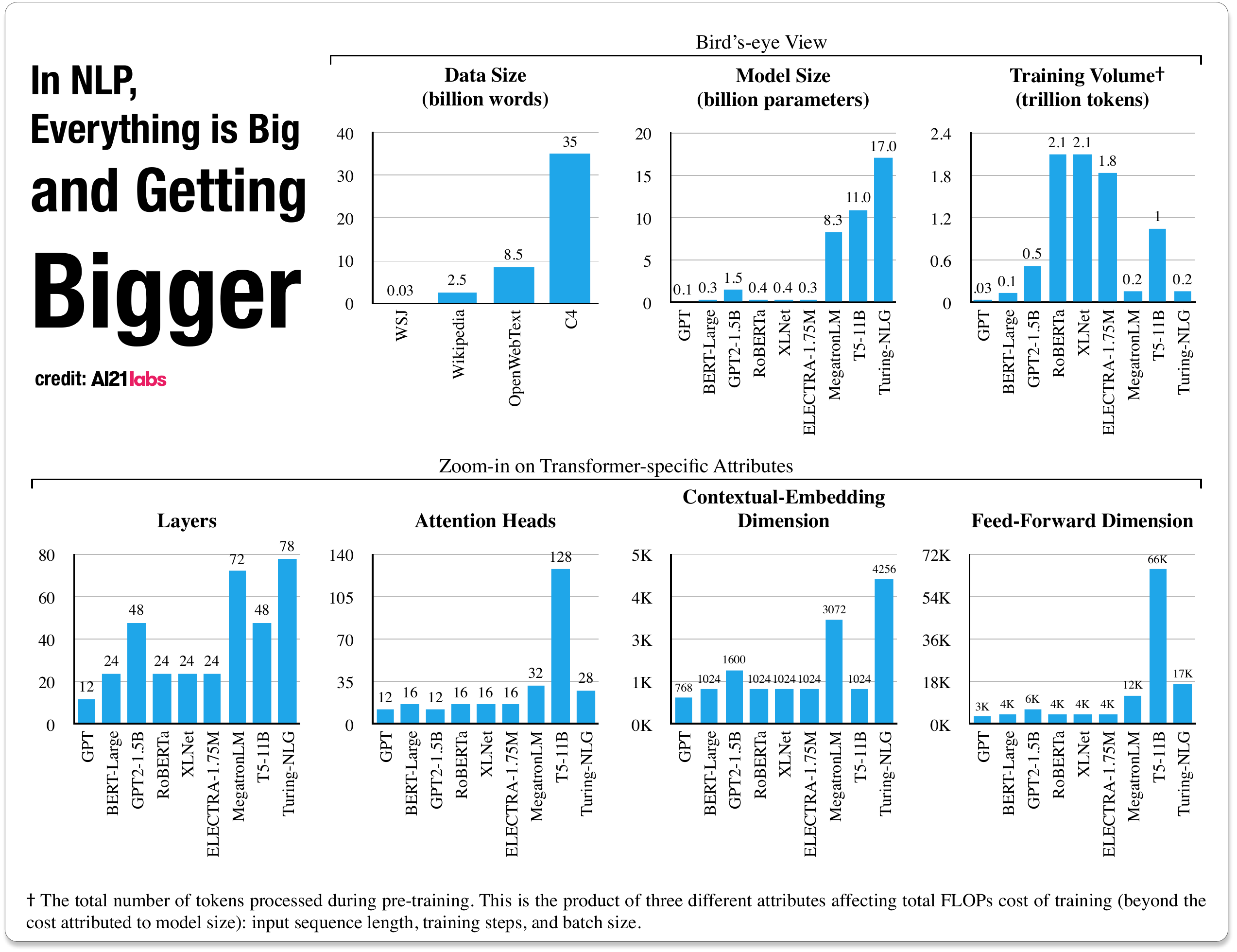}

\newpage
The exact ways in which these increases impact the number of FLOPs are subtle, and
depend on the specific training scheme and architecture. For example,
 fewer FLOPs are needed when training BERT-style models versus GPT-2~\citep{radford2019language} models with comparable model and data sizes, and training steps. Other training schemes can introduce additional factors that dictate cost; for example, the adversarial training scheme of
ELECTRA~\citep{Clark2020ELECTRA} uses an additional ``generator'' model during
training. This increases the relative per-step costs, but requires fewer steps,
thus reducing the overall costs. Despite these subtleties, however, it is clear
that all these growing numbers correlate with an overall trend towards a greater
number of FLOPs, which determine the bottom line.

On top of the above, there are also additional hidden costs, which are often overlooked. Each model
must be trained multiple times~--~this is in order to minimize random effects
(each run is inherently stochastic), and to search over a combinatorially large
hyper-parameter search space. This means there can be a large multiple over the
cost of a single training episode (although significant cost savings can be had
by conducting most of the experiments on the smaller models first, before training
the large models in the optimized configuration).

\section{The Future}
The reason the community has adopted the mega-scale, brute-force statistical approach
is that it works; it has yielded better performance than any alternative. And since
NLP has substantial economic value, no cost is too high in pursuit of good
performance. We do not see an end to the use of large NN models operating on massive
corpora, and one can imagine the
costs escalating further, as the community develops more elaborate architectures in
pursuit of more ambitious tasks. As you go from sentences to whole documents and beyond, you can
imagine the need for more dimensions per token, longer contexts, and potentially more layers.
Adding external knowledge sources, although potentially reducing the sole reliance on the network
(see below), could also contribute to expanding the size of the network in order to reflect the
external knowledge in the embedding space. Indeed, there is already discussion~\citep{SambaNovaModel}
of 100B-parameter models. That said, we see several factors that may help tame this
explosion and prevent things from getting out of hand. In increasing order of importance:
\begin{itemize}
    \item Further reduction of raw-compute prices due to increased competition. According to this (admittedly self-interested) blog post~\citep{AWSCosts}, the prices on AWS were reduced over 65 times since its launch in 2006, and by as much as 73\% between 2014 and 2017. We expect the same trend for AI-oriented compute offerings. 
    \item More efficient NN architectures, driven in part by economics and partly by environmental considerations. For example, the Reformer~\citep{Kitaev2020Reformer} architecture uses heuristics to reduce the complexity of the attention mechanism of transformers from quadratic to $O(n\log n)$. Similarly, ALBERT~\citep{Lan2020ALBERT} achieves better accuracy with fewer parameters by factorizing the embedding matrix and weight sharing across layers. We expect to see more of this.
    \item Ending the State-of-the-Art (SOTA) race. There is increasing recognition in the community that significant amount of compute is sunk into  reaching the top of leaderboards of the many challenge datasets, often involving many (in some reported cases, thousands) of runs, just so that one instance will luck into first place. Such overfitting is of course of little value, and we expect to see less of it. 
    \item Maxing out on useful data. There is just that much (useful) text that has been
          written, or that will be. At some point, we will have trained on
          \href{https://en.wikipedia.org/wiki/The_Library_of_Babel}{Borges' Universal Library}. 
    \item Useful as NNs are, there is a school of thought that holds that statistical ML is
          necessary but insufficient, and will get you just that far. Instead, the thinking
          goes, you need to incorporate structured knowledge and symbolic methods into the mix,
          and that in turn depends on brain rather than (only) brawn. This is a view we
          subscribe to at AI21 Labs (see~\citep{levine2019sensebert} as an example).
\end{itemize}

\medskip

\printbibliography

\appendix

\medskip

\section{NLP versus CV}\label{app:NLPvsCV}
With a few notable exceptions\footnote{There have been a few attempts to create
``mega-models'' for CV, e.g., FixResNet~\citep{FixResNet} has 830M parameters
and was trained on nearly a billion weakly-labeled images. However, the gains are
not as great compared to the added costs, and such approaches have not become the
norm just yet.}, you do not see in computer vision (CV) the large numbers and cost
escalations you do in NLP, and it is natural to ask why. We enter this discussion with some trepidation. While some of the folks at AI21 Labs have experience in CV, it is not our core competence as a company. Furthermore, some of the CV experts with whom we spoke did not have firm opinions here, and the opinions they did have did not always agree with each other. Still, since we have been asked this question we feel we should address it, but please treat the following more as a beginning of a discussion rather than definitive answers.\footnote{See also this article~\citep{NLPImageNet} for an interesting discussion circa 2018.} 

We believe that there are fundamentally two reasons why training CV models is cheaper than training NLP models:

\begin{itemize}
    \item \textbf{Images versus sentences.} Images are smooth and local, in that by and large the
          value of a pixel depends mostly on its close neighborhood and less so on other
          parts of the image, and furthermore the value does not change drastically from one
          pixel to its neighbor. Moreover, images are \emph{iconic}, by which we mean that "what you see is what you get"; an image of chair and a desk represents a chair and desk.  Language is very different. Words far apart can be coupled probabilistically, and language is \emph{compositional}; the way you string words together carries as much meaning as the semantic content of the words themselves. 
    \item \textbf{Object recognition versus~--~what?} The canonical problem in computer vision~--~object
          recognition/classification~--~is, while by no means trivial, relatively simple. It
          has no direct analog in NLP. One could argue that topic- or sentiment-analysis are
          somewhat analogous at the document level, and word-sense disambiguation is at the
          sentence level. But the analogy is weak, and neither of these plays the same central role
          that object recognition does in vision. Another telling analogy is between object
          identity in vision (is the person seen in this image the same as the person in this
          other image?) and noun-phrase co-reference in NLP (does ``the president'' refer to the
          same entity as ``Mr. Trump''?). Here the separation between vision and language is
          stark; object identity is close to being a solved problem, while co-reference is still
          unsolved. And this is leaving aside the issue that even once solved, co-reference on
          its own would not bring the same value that object recognition does in CV.
\end{itemize}
    
These differences manifest themselves in several ways, including these:

\begin{itemize}
\item \textbf{CNNs versus transformers.}
 CV problems lend themselves to
          Convolutional Neural Networks (CNNs), while the canonical NLP approach has centered
          around transformer models, which are inherently more expensive
          than CNNs. The different choice of architecture is directly related to
          the differences between images and sentences; the
          locality property matches with the local windows of convolutional layers, and
          smoothness with the sub-sampling operation in pooling layers. Since language does
          not enjoy these properties, we must use a more general, but less efficient,
          architecture such as the transformer.
    \item \textbf{Supervised versus semi-supervised versus self-supervised learning.} NLP and computer vision employ all of
          these learning regimes, but the balance is different. Unlike in computer vision,
          most of the training time of NLU models is devoted to self-supervised learning of
          language itself, and only a small portion is devoted to (supervised) fine-tuning
          of the model to solve a specific task. This is related to the inherent complexity
          of the structure of language and the nature of NLU and NLG tasks. Much larger
          datasets are needed in order to provide useful signal, and, just as bad, the tasks
          are inherently more ambiguous and the data is harder for people to annotate than
          in image classification; it is easier to answer the question ``Is that a person or
          a car'' than ``Does this sentence imply that sentence''. Furthermore, data
          augmentation is much more successful in vision than in NLP, and semi-supervised
          learning aided by data augmentation has led to many recent SOTA results in vision.
          In contrast, NLP has been driven toward purely self-supervised learning (``the
          NLP revolution will not be supervised!''). This in turn translates into larger
          training datasets compared to the supervised setting, as well as longer training cycles. 
\end{itemize}

Again, important caveats apply to all of the above. Even in object recognition, the larger context of the image can matter when determining what is depicted in a given image patch. Furthermore, object recognition is not the sole focus of CV, and more elaborate tasks, such as scene understanding~\citep{Geman3618}, certainly do not have the smooth, local properties mentioned (to use a famous example, object recognition techniques do not tell you the interesting part about an image depicting a piano dropping through the air and about to land on someone's head). As another example, in the area of image synthesis, which often requires accommodating complex logical, real-world constraints in the synthesized image, CNNs give way to inherently more expensive models such as GANs.  

Despite these important caveats, we feel the above analysis is fair, for two reasons. First, it is the case that among CV technologies, object recognition has brought the most commercial value to date, and CNNs have been the main driver behind its success. And second, more ambitious tasks such as scene understanding are getting close to NLP in being less well defined and less well solved. They also call for the same commonsense reasoning as does NLP, and thus are likely require the elaborate techniques -- and costs -- currently associated with NLP.  

\end{document}